\documentclass{article}

\usepackage[final]{neurips_2024}

\usepackage[utf8]{inputenc} % allow utf-8 input
\usepackage[T1]{fontenc}    % use 8-bit T1 fonts
\usepackage{hyperref}       % hyperlinks
\usepackage{url}            % simple URL typesetting
\usepackage{booktabs}       % professional-quality tables
\usepackage{amsfonts}       % blackboard math symbols
\usepackage{nicefrac}       % compact symbols for 1/2, etc.
\usepackage{microtype}      % microtypography
\usepackage{xcolor}         % colors
\usepackage{algorithm}
\usepackage{algorithmic}
\usepackage{graphicx}
\usepackage{amsmath}

\title{The EarlyBird Gets the WORM: Heuristically Accelerating EarlyBird Convergence}

\author{%
  Adithya Vasudev\\
  College of Computing\\
  Georgia Tech\\
  Atlanta, GA 30332 \\
  \texttt{avasudev8@gatech.edu} \\
}

\begin{document}

\maketitle

\begin{abstract}
The Lottery Ticket hypothesis proposes that ideal, sparse subnetworks, called lottery tickets, exist in untrained dense neural networks.
The Early Bird hypothesis proposes an efficient algorithm to find these winning lottery tickets in convolutional neural networks, using the novel concept of 
distance between subnetworks to detect convergence in the subnetworks of a model. 
However, this approach overlooks unchanging groups of unimportant neurons near the search's end. We propose WORM, a method that exploits 
these static groups by truncating their gradients, forcing the model to rely on other neurons. 
Experiments show WORM achieves faster ticket identification during training on convolutional neural networks, despite the additional computational overhead, when compared to EarlyBird Search.
Additionally, WORM-pruned models lose less accuracy during pruning and recover accuracy faster, improving the robustness of a given model.
Furthermore, WORM is also able to generalize the Early Bird hypothesis reasonably well to larger models, such as transformers, 
displaying its flexibility to adapt to more complex architectures.
\end{abstract}

\section{Introduction}

Deep Neural Networks (DNNs) have been shown to be a key component to autonomously intelligent systems in modern 
society, achieving impressive performance across a variety of domains and applications. However, these powerful models often 
require significant compute and energy resources, limiting their accessibility, scalability, and cost-effectiveness; 
even the smallest performant models require as many as 2 gigaflops per forward pass [2].
With the advent of large language models and their ubiquity across industries, this trend only seems to continue or worsen [3].

Recent trends in training efficient deep learning models have focused on a select few methods, namely pruning, quantization, and 
distillation [4]. Pruning, in particular has found widespread success as a method of reducing
the computational cost of training and deploying models due to its simplicity and scalability [5]. 
Traditional pruning involves fully training a model, 
then removing weights (pruning) and retraining to achieve target accuracy with fewer FLOPs.

While effective for cost reduction, traditional pruning suffers from slow execution. 
It requires full training, pruning, and then retraining to reach target accuracy. 
This increases training time, often negating efficiency gains due to retraining with fewer connections [1].

Recent research suggests that these ideal pruned networks, known as \textbf{subnetworks}, exist as a property of the initial architecture [5], and do
not require full training for discovery. These subnetworks emerge quickly, and can be estimated with high accuracy through efficient sampling 
methods. As a result, pruning algorithms have been developed to quickly detect the emergence of these subnetworks, and bypass most of the initial first training loop, 
inviting significant time savings [1].

However, one key issue with the current process is that these pruning algorithms fail to utilize all of the information contained within these estimated subnetworks, such as patterns on how they evolve. Patterns such as these may serve as great aids in understanding how the pruning process evolves and areas where it can be sped up. For example, the state-of-the-art approaches fail to consider 
patterns in the set of neurons that remain unaffected by pruning. Furthermore, due to the iterative sampling procedure [1] that these pruning methods use, they
have access to the evolution of the subnetwork over time. These pruning masks may reveal static patterns in the subnetworks, which if leveraged during training, could minimize retraining time and overall resource consumption.

This paper exploits this some of these properties by adding a heuristic to the search for these ideal pruned subnetworks, called WORM. 
More specifically, the WORM algorithm proposes the following three key ideas:
\begin{enumerate}
    \item Near search completion, the candidate mask stabilizes. We can use this mask to hint the model about upcoming pruning.
    \item Gradients encode learned information. Truncating gradients for masked connections effectively removes them from training.
    \item Early gradient masking forces the remaining network to learn lost information, reducing retraining time after pruning.
\end{enumerate}

Through experimental results and other findings backed by literature, this paper shows that all three key ideas hold. Furthermore, the
paper shows that this approach can also accelerate the detection of ideal subnetworks, thereby reducing total computation
time for training pruned subnetworks by as much as 5\%. In addition, this paper also shows that WORM is an ideal starting ground to discuss accelerated pruning approaches for much larger and more complex models, such as the transformers powering large language models.

\section{Background}

\subsection{The Lottery Ticket Hypothesis}
Proposed in 2019, the Lottery Ticket Hypothesis [5] was the first to point out that there exists a small subnetwork, 
called the winning ticket, that can achieve comparable test set accuracy when compared to the parent dense network.
It further claims that these efficient subnetworks are intrinsic to the architecture and potentially transferable across tasks.
However, the original paper and various other explorations into this domain only proposed 
that this winning ticket could be found through a process of iterative training and pruning.

\subsection{The Early-Bird Hypothesis}
Proposed and experimentally proved in 2022, the Early-Bird hypothesis [1] was shown to be a direct improvement to searching for the 
winning tickets theorized by the lottery ticket hypothesis. The hypothesis claims that lottery tickets can be found very early 
during the initial training process, and this hypothesis is validated by means of an algorithm to efficiently search for winning tickets. 
The paper performs this search using a novel concept called
\textbf{mask distance}, defined as the change in pruning masks across 
epochs. To speed up the computation of this metric, the paper focuses the computation of this distance only on batch normalization layers, due to their role in quantifying the importance of various channels 
across the convolutional neural network (CNN) [13][14]. This approach finds pruned subnetworks significantly faster than prior methods 
while achieving comparable or better accuracy.

\section{Properties of Early-Bird Search}
The key component to the Early-Bird hypothesis paper is the Early-Bird Search algorithm [1], which the original authors named EB Train.
EB Train relies on mask distance to identify stable pruned subnetworks. Masks are deemed stable, or converged, when the mask distance falls below a fixed value, called the \textbf{stopping point}. This value is computed by means 
of a maximal mask distance over a preset sliding window, with a window of 5 epochs often used in EB Train's experiments [1]. Due to the 
usage of a maximal mask distance in computing the convergence condition, this implies the existence of masks that become 
\textbf{stable enough} within the window.
As a result, we hypothesize that the mask distance often plateaus near the stopping point early in training. Thus, the majority of the epochs spent by the EB Train algorithm are used to reduce the mask distance by a very small value.

\subsection{Near-Convergence Behavior of Early-Bird Tickets}

\begin{figure}[htbp]
    \center{\includegraphics[scale=0.35]{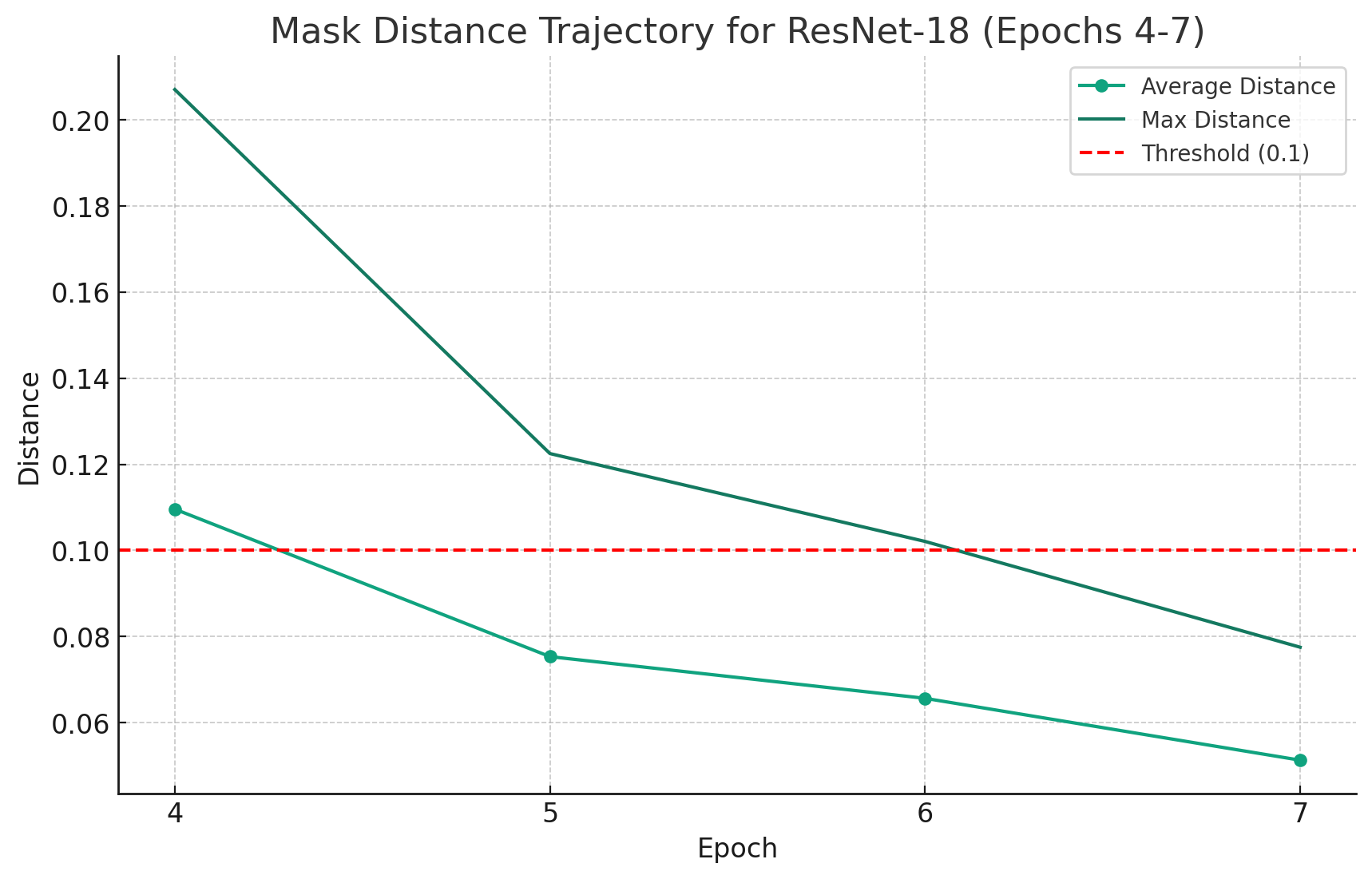}}
    \caption{Mask Distance Trajectory only for Full Sliding Window}
    \label{fig1}
\end{figure}

To validate these claims, we observe both the average mask
distance and the maximal mask distance over a training cycle
for a ResNet-18 model on the CIFAR-10 dataset [2][6]. Figure
\ref{fig1} shows the trajectories of both the average and maximal
distance, subject to the EB Train stopping condition of a
mask distance of 0.1 and a sliding window size of 5 epochs.
The reported results are an average over 7 runs with different
initialization. We begin analyzing the trajectory of the 
mask distance once the sliding window is full; this occurs at epoch 4.

As shown above, a distinctive elbow occurs for a maximal mask distance value that is substantially close to the stopping value of 0.1 (in this case, 
its within 0.02). This elbow is matched by a similar phenomenon occuring for the trajectory of the average distance. These observations
hold for various initalizations and across various runs. 
We can extrapolate these facts to thus show that near convergence, 
we observe stability in the masks, as indicated by the slowdown in convergence after the elbow. This stability suggests that we can use 
these masks to extract out patterns in the connections that will be pruned.

\section{Gradient Truncation}
One particularly successful method of helping models learn with external information is by modifying gradients based on a given 
condition. 
Two major techniques that have found widespread use in training are gradient clipping [9] and gradient normalization [7]. Gradient clipping in 
particular, has found widespread success in tackling the vanishing and exploding gradients problem, by penalizing individual gradients 
that exceed a given limit [8][9].

Here, we propose a method to inform the model about upcoming pruning heavily inspired by gradient clipping, called \textbf{gradient truncation}.
Instead of clipping the gradients to be no greater than a preset magnitude [9], we heavily truncate gradients regardless of their magnitude
for connections identified for pruning by the current mask, acting as a way to block information flow to these weights, 
forcing the model to focus on the remaining connections.

Formally, we can express gradient truncation using the following equation. 
Let $m(\theta)$ be a function outputting $1$ if $\theta$ represents a weight whose connection will be pruned as per the candidate mask, 
and $0$ otherwise. Then, we can define the truncated gradient function: $\nabla_{t}(\theta)$, as such:
\[
\nabla_{t}(\theta) = 
\begin{cases} 
r \nabla(\theta) & \text{if } m(\theta) = 1,\\ 
\nabla(\theta) & \text{if } m(\theta) = 0.
\end{cases}
\]
where $r$ is a parameter representing the proportion of the gradient to remain. We call this the \textbf{truncation parameter}.
If $r = 1$, then there is no truncation; we simply run the standard EB Train procedure. We apply this gradient truncation to each parameter 
that is pruned by the EB Train algorithm during its Early-Bird ticket sampling step. 

Because weights that are already marked as unimportant will be untouched, the probability 
that these weights still remain unimportant is high. As a result, this focuses the model to embed information 
in the remaining important connections, and thus, we hypothesize gradient truncation to lead to faster convergence [14]. Additionally, we expect the post-pruning accuracy to improve.
Since we are hinting to the model about which weights will get pruned, and eliminating additional information from being learned 
by these weights the model will be forced to learn this information using weights that are likely to not get pruned.
Nevertheless, we proceed to validate this claim with a series of experiments below.

\subsection{Impacts of Gradient Truncation on EB Train}
To experimentally analyze the performance of gradient truncation and its impact both during EB Train and retraining, 
we run multiple experiments using a convolutional neural network architecture. As previously mentioned, we insert the gradient truncation step 
into the ticket sampling step of EB Train: when the representative batch normalization layers are pruned, we step into 
each layer's gradients and clip them according to the previously mentioned rule. For the sake of analysis, 
we apply this truncation step at every single epoch.

Table \ref{table1} shows the results of various truncation parameters applied 
over the entirety of the EB Train algorithm, on the VGG-11 [11] model using the CIFAR-10 [6] dataset. Once again, we use the standard EB Train parameters 
of a sliding window of size 5, and a stopping distance of 0.1, as described by the original paper. Both algorithms use PyTorch's
unstructured $L_1$-norm magnitude pruning [12] as the algorithm of choice. We report the average of 
the results over 5 runs.

\begin{table}[htbp]
    \centering
    \caption{Impacts of Gradient Truncation on EB Train (VGG-11)}
    \label{table1}
    \begin{tabular}{|p{1.3cm}|p{1.3cm}|p{1.3cm}|p{1.3cm}|p{1.3cm}|}
    \hline
    Truncation Parameter ($r$) & Ticket Search Epoch Count & Post-Prune Accuracy & 90\% Restore Epochs & Total Execution Time \\
    \hline
    $r$ = 1 & 7 & 22\% & 1 & 237s \\
    \hline
    $r$ = 0.5 & 6 & 20\% & 2 & 239s \\
    \hline
    $r$ = 0.1 & 5 & 21\% & 3 & 237s \\
    \hline
    $r$ = 0.05 & 6 & 26\% & 2 & 239s \\
    \hline
    $r$ = 0.01 & 5 & 26\% & 2 & 213s \\
    \hline
    \end{tabular}
\end{table}

As expected, EB Train converged faster with gradient truncation, and both convergence time and post-pruning accuracy improved as truncation increased. 
An unexpected side effect that we observed was that retraining time also grew with more aggressive truncation. 
We attribute this fact to the model being restricted from learning key features early on due to the continuous
application of gradient truncation to the batch normalization layers. Since batch normalization 
layers encode the importance of particular channels for a CNN, clipping the gradients too early does not give the model a chance to learn important channels and other 
macroscopic features [1] [13] [14].

\section{The WORM Algorithm}
The previous results in training distortion, combined with frequent mask changes early in training, 
as illustrated by Figure \ref{fig1} suggest that a delayed start for gradient 
truncation may be more optimal, specifically at or around the elbow epoch.
Computing this elbow analytically is rather complex, and thus we take an experimental approach to computing this.
We propose a value, called the \textbf{trigger point},
defined as the point when the average mask distance dipping below a threshold. When the average mask distance falls below this trigger point, we begin applying gradient truncation.

To explain why average mask distance was selected as the representative measure, we naively began by letting the trigger point be a value that matched the convergence threshold (stopping point) for EB Train. Intuitively, to let truncation happen, we need to ensure that there exists at least one epoch where this representative quantity for the mask distance is below this trigger point. From the first experiment in Section 3, we see that the average mask distance is a great candidate, as it falls below the stopping point prior to when the maximal mask distance does, but it follows a similar trajectory to that of the maximal mask distance, meaning it doesn't fall too early. As a result, we apply gradient truncation for an ample number of epochs, but not early enough to eliminate the benefits of full training, as described in Section 4. 

To account for variance in performance of the underlying EB Search across individual runs, we add a small value $\delta$ to this trigger point. In our experiments, we let $\delta = 0.05$, 
to strongly ensure that enough rounds of gradient truncation would be applied to extract meaningful results. In practice, smaller 
values of $\delta$ are usable while still achieving similar results.

Once we achieve the elbow, we apply gradient truncation. As shown in the experimental results from Table \ref{table1}, lower 
truncation parameters can help accelerate convergence, while not significantly increasing the retraining time, as most of the increase in 
retraining time is incurred as a result of applying gradient truncation too early. To encourage extreme penalization of the 
pruned weights, we set $r$ to be an extremely small value; in our experiments, we let $r = 3 \times 10^{-3}$
We deem this approach as the \textbf{WORM} (Weighted Optimization for Robust Model training) algorithm.

\begin{algorithm}
    \caption{The WORM Algorithm}
    \begin{algorithmic}[1]
    \STATE \textbf{Initialize:} weights $W$, truncation parameter $r$, prune ratio $p$, early bird stopping point $s$, margin of error $\delta$, maximum number of epochs $N$, sliding window length $l$
    \STATE $r \leftarrow 1$
    \FOR{$i = 1$ \TO $N$}
        \STATE Train and update $W$ using a learning algorithm
        \STATE Prune the representative layers with a prune ratio $p$
        \STATE $\nabla(\theta) \leftarrow r \nabla_{t}(\theta)$ (application of Gradient Truncation)
        \STATE Compute the mask distance $d$ between the current mask and the previous $l$ masks, and update the sliding window
        \IF{$d_{\text{max}}$ $<$ $s$}
            \STATE Stop the algorithm and return the pruned mask as the candidate mask
        \ELSIF{$d_{\text{avg}}$ $<$ $s + \delta$}
            \STATE $r \leftarrow 3 \times 10^{-3}$
        \ENDIF
    \ENDFOR
    \end{algorithmic}
\end{algorithm}

\subsection{Comparisons with Naive Early-Bird Search}
To evaluate the WORM algorithm, we initialized two identical models with the same 
initialization and ran EB Train on one, and WORM on the other. Both approaches had a stopping parameter of 0.1, a sliding window of 5,
and a pruning rate of 0.5. Additionally, WORM had a margin of error $\delta = 0.05$ (leading to a trigger point of 0.15), and a truncation parameter of $r = 0.003$.
We evaluated both algorithms on two convolutional neural networks. First, both approaches were evaluated on ResNet-18 [2], as a baseline comparison against the model used in the original EarlyBird paper [1]. Then, to evaluate the ability for WORM to scale to larger convolutional neural networks, we evaluated both models on a variant of VGG-11 with batch normalization layers [11], which has roughly ten times the parameter count of ResNet-18. Both models were trained on the CIFAR-10 [6] dataset, using
 the stochastic gradient descent optimizer, and a learning rate of $\alpha = 0.01$, held statically, on a
singular NVIDIA H100 GPU. This set up was  run seven times, with all settings held identical except for the random seed used for initialization, which was changed across each run, but held the same for both algorithms. Timing was performed with the Python standard library. The average of the results were taken. The results for both ResNet-18 and VGG-11 are shown in Table \ref{table2} and Table \ref{table3}, respectively.

\begin{table}[htbp]
    \centering
    \caption{EB Train vs. WORM (ResNet-18)}
    \begin{tabular}{|p{2cm}|p{2cm}|p{2cm}|p{2cm}|p{2cm}|p{2cm}|}
    \hline
    Algorithm & Ticket Search Epoch Count & Pre-Pruned Acc. & Retrain Acc. after One Epoch & 90\% Restoration Epochs & Total Time \\ \hline
    EB Train & 6 & 81.67\% & 84.67\% & 3.3 & 264s \\ \hline
    WORM & 5 & 80.67\% & 82.67\% & 4 & 262s \\ \hline\end{tabular}
    \label{table2}
\end{table}

\begin{table}[htbp]
    \centering
    \caption{EB Train vs. WORM (VGG-11)}
    \begin{tabular}{|p{2cm}|p{2cm}|p{2cm}|p{2cm}|p{2cm}|p{2cm}|}
    \hline
    Algorithm & Ticket Search Epoch Count & Pre-Pruned Acc. & Retrain Acc. after One Epoch & 90\% Restoration Epochs & Total Time \\ \hline
    EB Train & 6.7 & 95.33\% & 90.67\% & 1 & 229s \\ \hline
    WORM & 5.7 & 92.33\% & 89.67\% & 1.7 & 222s \\ \hline 
    \end{tabular}
    \label{table3}
\end{table}

As seen in the two tables above, WORM consistently found winning tickets faster than EB Train despite needing more computations, as 
gradient masking doubles the FLOP count of the sampling step in EB Train.
It achieves similar accuracy and recovers accuracy quickly after pruning, even surpassing EB Train in some cases.
While WORM requires slightly longer retraining, likely due to the fact that that the pre-pruned accuracy on average
was lower due to the fewer number of epochs the model had to train, combined with the fact that the model training was being punished 
by truncating the gradients, its efficiency gains in finding winning tickets outweigh this cost. 

Table \ref{table3} shows 
that WORM was able to beat EB Train on the amount of accuracy recovered in 1 epoch consistently, while \ref{table2} showed that WORM was able 
to stay within 1\% of the accuracy recovery of EB Train. On average, WORM was able to perform on average 1\% faster on ResNet-18, and 4\% faster on VGG-11. As a result, WORM is able to further improve on EB Train's core goal of improving the 
efficiency of training the large and dense model while also searching for potential winning tickets, while at the minimum providing 
comparable retraining efficiency to EB Train, occasionally beating it. 

Table \ref{table3} also showed that WORM was remarkably robust in terms of scalability, holding the same gains that EB Train had when applied to larger models, despite the increased overhead in gradient truncation computation due to the larger model. This shows that WORM scales about as well as EB Train, and can keep its performance improvements (or even exceed them) on these larger models. 

\subsection{Limitations}

Despite these improvements, it remains important to note that the WORM algorithm is theoretically more computationally expensive in terms of floating-point operations, as it requires two sweeps through the pruned representative layers, versus one in the EB Train algorithm. Since the concept of representative layers hinges on the key assumption that these layers are important despite their relatively rare distribution across the model, it's possible that for some architectures, the representative layer may be significantly more common, leading to degeneration of the performance improvements due to the increased overhead. Nevertheless, due to the assumption holding for the tested models, it clearly shows that WORM is a direct improvement to finding early bird tickets on convolutional neural networks.

\section{Extensions to Transformers}

Over the past few years, the transformer [3] architecture has exploded in popularity due to its impressive performance in language 
modeling. However, these models come at the cost of extremely high compute costs for both training and inference, due to their sheer 
size, layer counts, and layer complexity [16].
As a result, finding similar methods for reducing the computational complexity for these larger models is an area 
of open research. Recent works have shown that lottery ticket
pruning can improve efficiency, however, existing methods differ from EB Train's approach, as they typically apply post-training approaches, as opposed to EB Train's prune-while-train approach [21].

\subsection{Extending EB Train to Transformers}
EB Train uses concept of representative layers to observe how 
the pruned masks would evolve over time without sampling the entire model [1]. For modern convolutional neural networks, this layer is often the batch normalization layer, as
it represents the importance of the various channels in the CNN [13] [14]. Thus, the first step to extending EB Train is to determine which layer 
class will serve as the representative layer.

The core feature of a transformer is the concept of multi-head attention, which attends to previous (and future tokens, if its bidirectional). This 
layer effectively acts as an importance mapping of each token relative to each other [15]. Thus, this module, and more specifically, the linear 
layers representing the key, query, and value matrices, will serve as the representative layers for EB Train.

We first run an experiment to determine the feasibility of running EB Train, using multi-head attention as the representative layers. 
We execute three runs of EB Train on the BERT [16] model, using the GLUE [17] dataset to evaluate accuracy. We use a pretrained instance of BERT, 
specifically one using a Sentence Classification head [20], and finetune the model on a single NVIDIA H100, using three GLUE tasks: SST2, RTE, and MRPC. Using a stopping 
parameter of 0.01, a pruning ratio of 0.5, and a sliding window of 3 epochs. The lower mask distance was set due to the fact that the model 
was pretrained. The results for this are shown in the table below:

\begin{table}[h!]
    \centering
    \label{table4}
    \caption{Performance of EB Train on GLUE Tasks}
    \begin{tabular}{|p{6cm}|p{2cm}|p{2cm}|p{2cm}|}
    \hline
    \textbf{EB Train}                           & \textbf{SST}    & \textbf{MRPC}   & \textbf{RTE}    \\ \hline
    \textbf{SOTA Baseline}                      & 93.5\%          & 88.9\%          & 66.4\%          \\ \hline
    \textbf{Pre-Pruned Accuracy}                & 89.33\%         & 81.86\%         & 64.98\%         \\ \hline
    \textbf{Epochs to Prune}                    & 4               & 3               & 3               \\ \hline
    \textbf{Pruned Accuracy}                    & 86.35\%         & 78.67\%         & 50.90\%         \\ \hline
    \textbf{Accuracy after 1 Epoch of Retraining} & 88.76\%       & 83.57\%         & 62.09\%         \\ \hline
    \textbf{Accuracy Gain after 1 Epoch}        & -0.56\%         & +1.71\%         & -2.89\%         \\ \hline
    \textbf{Time to Prune and Retrain}          & 770s            & 40s             & 41s             \\ \hline
    \end{tabular}
\end{table}

We find that EB Train is highly effective for finetuning, finding convergent masks in as few as three epochs on certain tasks. 
As a further experiment, we also experiment with only pruning the query and key linear layers of the multi-head attention module, as 
a time saving mechanism, and explore its performance in both convergence, retraining, and accuracy. We run the results on the RTE task, 
for the sake of simplicity. 

\begin{table}[h!]
    \centering
    \caption{EB Train w/o Value Masking}
    \label{table5}
    \begin{tabular}{|p{6cm}|p{2cm}|}
    \hline
    \textbf{EB Train} & \textbf{RTE} \\ \hline
    Pre-Pruned Accuracy & 66.1\% \\ \hline
    Epochs to Prune & 3 \\ \hline
    Pruned Accuracy & 51.3\% \\ \hline
    Accuracy after 1 Epoch of Retraining & 65.03\% \\ \hline
    Accuracy Gain after 1 Epoch & -1.37\% \\ \hline
    Time to Prune and Retrain & 38s \\ \hline
    \end{tabular}
\end{table}

We find that not pruning the value layer still allows for similar performance, while speeding up the computation time and reducing 
the total number of FLOPs needed for EB Train. Repeated trials also showed more predictable convergence behavior. As a result, future experiments using EB Train on transformers will be conducted 
using only the query and key linear layers as the representative layers.

\subsection{Extending WORM to Transformers}
We turn our attention to adapting the WORM approach to Transformers. Similar to EB Train, 
we select the multi-head attention module, and specifically the query and key linear layers, as the representative layers for 
pruning. With the stopping parameter being 0.01 for standard EB Train, we let $\delta = 0.005$ for WORM, yielding a 
trigger point of 0.015. We also use a truncation rate of $r = 0.003$, as we did in the experiments for ResNet-18 and VGG-11. 
Once again, we evaluate WORM and EB Train using a pretrained instance of BERT [16] with a Sentence Classification [20] head.
We finetune the model on a single NVIDIA H100, using three GLUE [17] tasks: SST2, RTE, and MRPC. The results for this experiment are in the 
table below, averaged over three runs.

\begin{table}[h!]
    \centering
    \caption{Performance of WORM on GLUE Tasks}
    \label{table6}
    \begin{tabular}{|l|c|c|c|}
    \hline
    \textbf{WORM} & \textbf{SST} & \textbf{MRPC} & \textbf{RTE} \\ \hline
    Pre-Pruned Accuracy & 88.53\% & 82.59\% & 65.34\% \\ \hline
    Epochs to Prune & 4 & 4 & 4 \\ \hline
    Pruned Accuracy & 88.30\% & 76.90\% & 52.70\% \\ \hline
    Accuracy after 1 Epoch of Retraining & 88.81\% & 83.82\% & 62.81\% \\ \hline
    Accuracy Gain after 1 Epoch & +0.28\% & +1.23\% & -2.81\% \\ \hline
    Time to Prune and Retrain & 1072s & 65s & 51s \\ \hline
    \end{tabular}
\end{table}

As shown above, WORM achieved smaller accuracy drops than EB Train but with less retraining gain and lower speedup compared to CNNs. 
Interestingly, WORM slowed down EB Train convergence for transformers. This might be due to distorting pre-trained information 
by truncating gradients. The loss plot of a run of WORM on the SST-2 task reveals this fact to be true, as the loss curve 
flattens very prematurely when WORM is applied. However, more experimental analysis is needed to validate this claim.

\begin{figure}[htbp]
    \center{\includegraphics[scale=0.35]{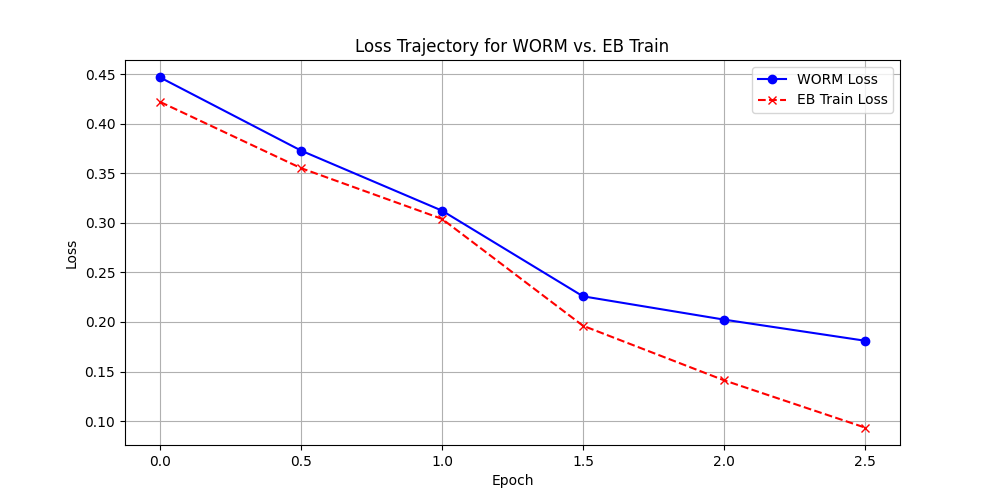}}
    \caption{Loss Trajectory for WORM vs. EB Train}
    \label{fig2}
\end{figure}

These results, taken collectively, show that directly applying WORM to transformers, while a good foundation for more intricate algorithms, does not retain ALL of the benefits it presents with convolutional neural networks. Although the experiments demonstrate that WORM enables models to retain more information even with high pruning rates (as shown by the low drops in accuracy with pruning), the additional computation overhead needed for the algorithm, combined with the matched performance in accuracy recovery may imply that this algorithm will not scale as well as it does for convolutional neural networks.

\section{Conclusion}

EB Train efficiently finds winning tickets but does not leverage all of the information gained during the search phase. 
WORM leverages this information by applying Gradient Truncation, forcing the model to learn mostly using important connections 
before pruning. This improves post-pruning accuracy, retraining speed, and winning ticket convergence for CNNs. 
While slower for transformers, WORM remains a promising first step for efficient transformer pruning methods.

In addition, despite these highly promising findings, testing WORM's capabilities across an even more diverse set of architectures, such as RNNs and larger transformers, is quite necessary, but due to time and compute constraints, such a diverse variety of experiments was not reasonably feasible. Nevertheless, due to their performance on CNNs, WORM still remains both a robust improvement over the naive EB Train algorithm, and a great baseline for designing more complex algorithms that are more computationally efficient.

In the future, we hope to further investigate more complex approaches for hinting pretrained transformers, as WORM typically incites a noticeable increase in compute time, for marginally improved 
robustness. Finally, investigating how WORM interacts with more complex pruning strategies, such as gradient pruning, 
structured pruning, and undecayed pruning, can provide insight into how the behavior of prune masks generalize near the convergence boundary.

\section*{Acknowledgments}

Thank you to Dr. Celine Lin and the three teaching assistants (Yonggan Fu, Haoran You, and Zhongzhi Yu) for the course CS 8803: Efficient Machine Learning 
at Georgia Tech, for both introducing me to the idea of Early Bird pruning, for motivating the core idea 
behind this project, and for providing feedback throughout my investigation. In addition, thank you to PACE @ Georgia Tech for providing the compute resources, 
to train the models needed for the project. 

\section*{References}

{
\small

[1] Haoran You, et al. "Drawing early-bird tickets: Towards more efficient training of deep networks," in CoRR, vol. abs/1909.11957, 2019.

[2] Kaiming He, et al. Deep Residual Learning for Image Recognition, in CoRR, vol. abs/1512.03385, 2015.

[3] Ashish Vaswani, et al. Attention Is All You Need, in CoRR, vol. abs/1706.03762, 2017.

[4] Geoffrey Hinton, Oriol Vinyal, Jeff Dean. Distilling the Knowledge in a Neural Network. 2015.

[5] Jonathan Frankle, Michael Carbin. The Lottery Ticket Hypothesis: Finding Sparse, Trainable Neural Networks. 2019.

[6] Alex Krizhevsky. Learning Multiple Layers of Features from Tiny Images. 2009.

[7] Yang Zhao, Hao Zhang, Xiuyuan Hu. Penalizing Gradient Norm for Efficiently Improving Generalization in Deep Learning. 2022.

[8] Jingzhao Zhang, Tianxing He, Suvrit Sra, Ali Jadbabaie. Why gradient clipping accelerates training: A theoretical justification for adaptivity. 2020.

[9] Razvan Pascanu, Tomas Mikolov, Yoshua Bengio. On the difficulty of training Recurrent Neural Networks. 2013.

[10] Omer Elkabetz, Nadav Cohen. Continuous vs. Discrete Optimization of Deep Neural Networks, in Advances in Neural Information Processing Systems. 2021.

[11] Karen Simonyan, Andrew Zisserman. Very Deep Convolutional Networks for Large-Scale Image Recognition. 2015.

[12] Adam Paszke, et al. PyTorch: An Imperative Style, High-Performance Deep Learning Library. 2019.

[13] Zhuang Liu, Jianguo Li, Zhiqiang Shen, Gao Huang, Shoumeng Yan, Changshui Zhang. Learning Efficient Convolutional Networks through Network Slimming. 2017.

[14] Selvaraju, R., Cogswell, M., Das, A., Vedantam, R., Parikh, D., Batra, D. Grad-CAM: Visual Explanations from Deep Networks via Gradient-Based Localization. International Journal of Computer Vision, 128(2), 336–359. 2019.

[15] Ma, W., Zhang, K., Lou, R., Wang, L., Vosoughi, S. Contributions of Transformer Attention Heads in Multi- and Cross-lingual Tasks. In Proceedings of the 59th Annual Meeting of the Association for Computational Linguistics and the 11th International Joint Conference on Natural Language Processing (Volume 1: Long Papers). Association for Computational Linguistics. 2021.

[16] Jacob Devlin, Ming-Wei Chang, Kenton Lee, Kristina Toutanova. BERT: Pre-training of Deep Bidirectional Transformers for Language Understanding. 2019.

[17] Alex Wang, Amanpreet Singh, Julian Michael, Felix Hill, Omer Levy, Samuel R. Bowman. GLUE: A Multi-Task Benchmark and Analysis Platform for Natural Language Understanding. 2019.

[19] Aidan Good, Jiaqi Lin, Hannah Sieg, Mikey Ferguson, Xin Yu, Shandian Zhe, Jerzy Wieczorek, Thiago Serra. Recall Distortion in Neural Network Pruning and the Undecayed Pruning Algorithm. 2022.

[20] Thomas Wolf, et al. Transformers: State-of-the-Art Natural Language Processing. In Proceedings of the 2020 Conference on Empirical Methods in Natural Language Processing: System Demonstrations (pp. 38–45). Association for Computational Linguistics. 2020.

[21] Xiaohan Chen, Yu Cheng, Shuohang Wang, Zhe Gan, Zhangyang Wang, Jingjing Liu. EarlyBERT: Efficient BERT Training via Early-bird Lottery Tickets. 2021.

\end{document}